\documentclass{article}

\usepackage[final]{corl_2019} 
\usepackage{amsfonts}

\usepackage{amsmath,amsfonts,bm}
\usepackage{subcaption}

\usepackage[utf8]{inputenc} 
\usepackage[T1]{fontenc}    
\usepackage{booktabs}       
\usepackage{amsfonts}       
\usepackage{nicefrac}       
\usepackage{microtype}      
\usepackage{wrapfig}
\usepackage{graphicx}
\usepackage{bm}
\usepackage{amsmath}
\usepackage{amssymb}
\usepackage{color}
\usepackage{algorithm}
\usepackage{algpseudocode}
\usepackage{amsmath,amsthm}
\newtheoremstyle{problemstyle}  
        {3pt}                                               
        {3pt}                                               
        {\normalfont}                               
        {}                                                  
        {\bfseries\itshape}                 
        {\normalfont\bfseries:}         
        {.5em}                                          
        {}                                                  
\theoremstyle{problemstyle}
\usepackage[utf8]{inputenc}
\usepackage[english]{babel}
\usepackage{appendix}
 
\newtheorem{theorem}{Theorem}

\title{Graph Policy Gradients for Large Scale Robot Control}
\author{
  Arbaaz Khan, Ekaterina Tolstaya, Alejandro Ribeiro, Vijay Kumar\\
  GRASP Lab, University of Pennsylvania\\
}

\begin{document}
\maketitle


\begin{abstract}
In this paper, the problem of learning policies to control a large number of homogeneous robots is considered. To this end, we propose a new algorithm we call Graph Policy Gradients (GPG) that exploits the underlying graph symmetry among the robots. The curse of dimensionality one encounters when working with a large number of robots is mitigated by employing a graph convolutional neural (GCN) network to parametrize policies for the robots. The GCN reduces the dimensionality of the problem by learning filters that aggregate information among robots locally, similar to how a convolutional neural network is able to learn local features in an image. Through experiments on formation flying, we show that our proposed method is able to scale better than existing reinforcement methods that employ fully connected networks. More importantly, we show that by using our locally learned filters we are able to zero-shot transfer policies trained on just three robots to over hundred robots. A video demonstrating our results can be found ~\href{https://www.youtube.com/watch?v=RefiX9UCCw8}{\textbf{here}.} Code for this paper can be found ~\href{https://github.com/arbaazkhan2/gpg_labeled}{\textbf{here}.}
\end{abstract}



\section{Introduction}

In the recent past, deep learning has proved to be an extremely valuable tool for robotics. Harnessing the power of deep neural networks has emerged as a successful approach to designing policies that map sensor inputs to control outputs for complex tasks. These include, but are not limited to, learning to play video games~\cite{dqn,mnih2016asynchronous}, learning complex control policies for robot tasks~\cite{visuomotor} and learning to plan with only sensor information~\cite{pathak2017curiosity,macn,gupta2017cognitive}. While these works are impressive, they often concern themselves with controlling a single entity. However, in  many real world applications, especially in fields like robotics there exists a need to control multiple robots interact with each other in co-operative or competitive settings. Examples include warehouse management with teams of robots~\cite{enright2011optimization}, multi-robot furniture assembly~\cite{knepper2013ikeabot}, and concurrent control and communication for teams of robots~\cite{2017Transactions_Stephan}. In such a scenario, as the number of robots increases, the dimensionality of the input space and the control space both increase making it much harder to learn meaningful policies. 

This paper looks to tackle the problem of learning individual control policies by exploiting the underlying graph structure among the robots. We start with the hypothesis that the difficulty of learning scalable policies for multiple robots can be attributed to two key issues: dimensionality and partial information. Consider an environment with $\mathbf{N}$ robots (This work uses bold font when talking about a collection of items, vectors and matrices).
Each robot receives partial observations of the environment. In order for a robot to learn a meaningful control policy, it must interact with some subset of all agents, $\mathbf{n} \subset \mathbf{N}$. Finding the right subset of neighbors to learn from is in itself a challenging problem. 
Further, in order to ensure that the method scales, as the number of robots increase, one needs to ensure that the cardinality of the subset of neighbors $|\mathbf{n}|$, remains fixed or grows very slowly. 
\begin{figure}[t!]
  \centering
  \includegraphics[scale=0.3]{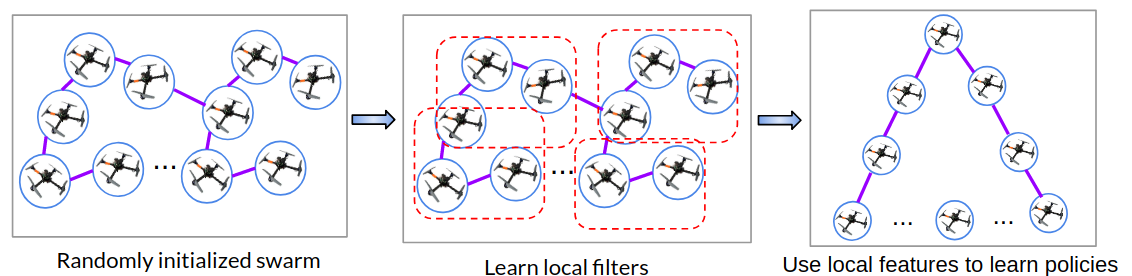}
  \caption{\textbf{Graph Policy Gradients.} Robots are randomly initialized and, based on some user set thresholds, a graph is defined. Information from K-hop neighbors is aggregated at each node by learning local filters. These local features are then used to learn policies to produce desired behavior.\label{fig:mainfig}}
  \vspace{-0.2cm}
\end{figure}
To solve these problems for large scale multi-robot control, we draw inspiration from convolutional neural networks (CNNs). CNNs consist of sequentially composed layers where each layer comprises of banks of linear time invariant filters to extract local features along with pooling operations to reduce the dimensionality. 
Convolutions regularize the linear transform to exploit the underlying structure of the data and hence constrain the search space for the linear transform that minimizes the cost function. However, CNNs cannot be directly applied to irregular data elements that have arbitrary pairwise relationships defined by an underlying graph. To overcome this limitation, there has been the advent of a new architecture called graph convolutional networks (GCNs) ~\cite{kipf2016semi,gama2018convolutional,wu2019comprehensive}. Similar to CNNs, GCNs consist of sequentially composed layers that regularize the linear
transform in each layer to be a graph convolution with a bank of graph filters and the weights of the filter are learned by minimizing some cost function. 

When controlling a large swarm of robots operating in a Euclidean space $\mathbb{R}^\mathfrak{n}$, the underlying graph can be defined as $\mathcal{G} = (\mathbf{V},\mathbf{E})$ where $\mathbf{V}$ is the set of nodes and $\mathbf{E}$ is the set of edges.
For example, we define each robot to be a node and add an edge between robots if they are are less than $\epsilon$ distance apart. This graph acts as a support for the data vector $\textbf{x}=[\mathbf{x}_1,\ldots,\mathbf{x}_N]^\top$ where $\mathbf{x}_n$ is the state representation of robot $n$. Now, our GCN exploits this graph $\mathcal{G}$ and at each node aggregates information from its neighbors (See Section \ref{sec:gcn}). This information is propagated forward to the next layer through a non linear transformation. The output of the final layer is given by $\Pi = [\pi_1,\ldots,\pi_N]$,  where $\pi_1,\ldots,\pi_N$ are independent policies for the robots. Similar to standard reinforcement learning, we execute these policies in the environment, collect a centralized reward and use policy gradients ~\cite{sutton1998reinforcement} to update the weights of policy network. We call this algorithm Graph Policy Gradients (GPG) (See Fig.~\ref{fig:mainfig}). 

One possible concern with our proposed algorithm is that, when working with on-policy methods for a large number of robots, it is highly likely that, due to exploration/entropy the graph might change during training. Such a setting can result in an explosion in the number of possible graphs that one needs to learn over as the number of robots increases. We show that in light of this, our choice of GCNs for policy parametrization is well motivated because it can be shown that graph convolutions are permutation equivariant, i.e if one were to reorder the node ordering of the graph and the corresponding graph signal, then the output of the graph convolution does not change ~\cite{gama2019stability}. 
This is important because it reduces the dimensionality of the problem and helps training converge faster when training with many robots.  

To demonstrate the efficacy of our proposed algorithm, we perform experiments on simulated formation flying with robots. Designing controllers and trajectories for formation flying is an important problem in multi-robot literature ~\cite{turpin2012trajectory,turpin2012decentralized} and is in fact a good test bed for other multi-robot control problems~\cite{kumar2012opportunities}. In our experiments, it is shown that GPG is able to converge as the number of robots  are increased in comparison to state of the art on-policy reinforcement learning algorithms that employ fully connected networks to parametrize the policy. We also show that our graph filters are able to learn valid local features by training them on a small number of robots and transferring the behavior to a very large number of robots. For example, we train a GCN for three robots to maintain formation, avoid collisions and follow their trajectories. Then, we initialize a swarm of many robots and use this same graph filter over the entire graph to generate desired behavior. We show that the desired formation flying behavior is achieved without updating the weights for the larger swarm thus achieving zero-shot transfer. Lastly, the ability of GPG to adapt to more complex dynamics and control is demonstrated. 

\section{Methodology}


\subsection{Preliminaries}
We pose learning the controller for a large number of robots as a policy learning problem in a collaborative Markov team~\cite{littman1994markov}. The team is composed of $\mathbf{N}$ robots generically indexed by $n$ which at any given point in time $t$ occupy a position $\mathbf{x}_{nt}\in\mathcal{X}$ in configuration space and must choose an action $\mathbf{a}_{nt}\in\mathcal{A}$ in action space. 
The team and environment are assumed to be Markovian. Thus, if one were to collect the robot configurations in the vector $\textbf{x}_{t} := [\mathbf{x}_{1t}, \ldots, \mathbf{x}_{Nt}]^{\top}$ and actions in the vector $\textbf{a}_{t}:=[\mathbf{a}_{1t}, \ldots, \mathbf{a}_{Nt}]$ then, the evolution of the system is completely determined by the conditional transition probability $p \left(\textbf{x}_{t+1}| \textbf{x}_t, \textbf{a}_t\right)$. The transition dynamics are also assumed to be the same for all agents, so that if we swap two of them in configuration and action space we expect to see the same statistical evolution. This is a natural consequence when, the robotic swarm is assumed to be composed of homogeneous robots. In this work, there exists no external communication between robots. Instead, we define a graph $\mathcal{G}$ that encodes information about each robots neighbors. Robot $n$ can directly communicate only with its one hop neighbors. Communication with neighbors that are further away is possible only indirectly. Each robot has access to its own states $\mathbf{x}_{nt}$ and $\mathcal{G}$. In this paper, we do not design which nodes each robot should talk to or what each robot must communicate. Each robot must learn a policy (probability distribution) from which the robot samples actions; $\mathbf{a}_{nt} := \pi_{n}(\mathbf{a}_{nt}|\mathbf{x}_{nt},\mathcal{G})$. Let $\Pi=  [\pi_{1},\ldots,\pi_{n}]$. As robots operate in the environment, they collect a \textit{centralized} global reward $r_t$. Collapsing everything, we are interested in computing $\textbf{a}_t := \Pi (\textbf{a}_t|\textbf{x}_t,\mathcal{G})$ such that the expected sum of rewards over some time horizon $T$ is maximized: 
\begin{equation}
\label{eq:overallloss}
\sum_{n=1}^N \max_{\theta}  \mathbb{E}_{\Pi}\bigg[\sum_{t}^T r_t\bigg]  \enspace
\end{equation}
where $\theta$ are the parameters of $\Pi$. At a high level, the global reward encodes desired behavior for the swarm. As the number of robots increase, the dimensionality of $\textbf{a}_t$ and $\textbf{x}_t$ increases too making the problem of learning $\Pi(\textbf{a}_t|\textbf{x}_t)$ non-trivial. Our proposition here to instead learn $\Pi(\textbf{a}|\textbf{x},\mathcal{G})$ only compounds the difficulty of the problem as the size of the graph grows exponentially as the number of robots increase ($N^2$ for $N$ robots). In the next section, we discuss using a graph convolutional network as a parametrization for $\Pi$ to overcome this problem by extracting local information from the graph structure.
\subsection{Graph Convolutional Networks}\label{sec:gcn}
\begin{figure*}[t!]
    \centering
    \begin{subfigure}{0.200\textwidth}
        \centering
        \includegraphics[width=\textwidth]{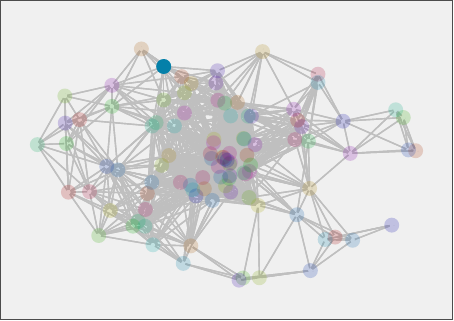}
        \caption{$\mathbf{y}^0$}
        \label{Agg_t0}
    \end{subfigure}
    \hfill
    \begin{subfigure}{0.200\textwidth}
        \centering
        \includegraphics[width=\textwidth]{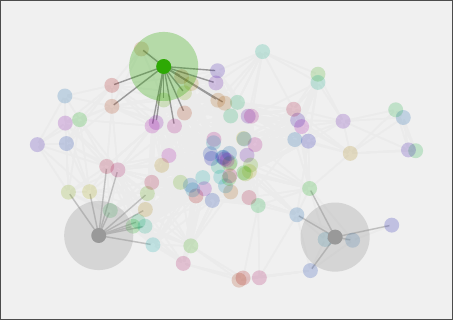}
        \caption{$\mathbf{y}^1$}
        \label{Agg_t1}
    \end{subfigure}
    \hfill
    \begin{subfigure}{0.200\textwidth}
        \centering
        \includegraphics[width=\textwidth]{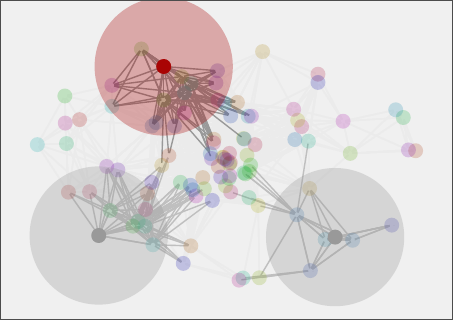}
        \caption{$\mathbf{y}^2$}
        \label{Agg_t2}
    \end{subfigure}
    \hfill
    \begin{subfigure}{0.2\textwidth}
        \centering
        \includegraphics[width=\textwidth]{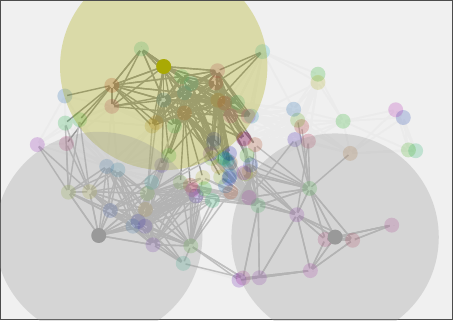}
        \caption{$\mathbf{y}^3$}
        \label{Agg_t3}
    \end{subfigure}
\caption{\textbf{Graph Convolutional Networks}. GCNs aggregate information between nodes and their neighbors. For each $k$-hop neighborhood (illustrated by the increasing disks), record $\mathbf{y}_{kn}$  (Eq. \ref{eq:graph_signal}) to build $\mathbf{z}$ which exhibits a regular structure (Eq. \ref{eq:finalapproxform}). \textbf{a)} The value at each node when initialized and at the \textbf{b)} one-hop neighborhood. \textbf{c)} two-hop neighborhood. \textbf{d)} three-hop neighborhood.}
	\label{fig_aggregation}
\vspace{-0.2cm}
\end{figure*}

Consider a graph $\mathcal{G}=(\mathbf{V},\mathbf{E})$ described by a set of $\mathbf{N}$ nodes denoted $\mathbf{V}$, and a set of edges denoted $\mathbf{E} \subseteq \mathbf{V} \times \mathbf{V}$.  This graph is considered as the support for a data signal $\mathbf{x} = [x_1,\ldots,x_N]^\top$ where the value $x_n$ is assigned to node $n$. The relation between $\mathbf{x}$ and $\mathcal{G}$ is given by a a matrix $\mathbf{S}$ called the graph shift operator. The elements of $\mathbf{S}$ given as $s_{ij}$ respect the sparsity of the graph, i.e $s_{ij} = 0$, $\forall$ $i\neq j \text{ and } (i,j) \notin \mathbf{E}$. Valid examples for $\mathbf{S}$ are the adjacency matrix, the graph laplacian, and the random walk matrix. In this paper, we consider the normalized graph Laplacian similar to  ~\cite{kipf2016semi}: 
\begin{equation}\label{eq:graph_laplac}
    \mathbf{S} = \mathbf{I}_N - \mathbf{D}^{-\frac{1}{2}}\mathbf{AD}^{-\frac{1}{2}}
\end{equation}
A key property of $\mathbf{S}$ is that it is assumed symmetric, with decomposition $\mathbf{S} = \mathbf{V} \Lambda \mathbf{V}^\top$ where $\mathbf{V}$ is the eigenvector matrix and $\Lambda$ is the eigenvalue matrix of $\mathbf{S}$.  
$\mathbf{S}$ defines a map $\mathbf{y} = \mathbf{S}\mathbf{x}$ between graph signals that represents local exchange of information between a node and its one-hop neighbors. More concretely, if the set of neighbors of node $n$ is given by $\mathfrak{B}_n$ then : 
\begin{equation}\label{eq:graph_signal}
    y_n = [\mathbf{S}\mathbf{x}]_n =\sum_{j=n,j\in \mathfrak{B}_n}s_{nj}x_n
\end{equation}
Eq. \ref{eq:graph_signal} performs a simple aggregation of data at node $n$ from its neighbors that are one-hop away. The aggregation of data at all nodes in the graph is denoted $\mathbf{y}=[y_1,\ldots,y_N]$. By repeating this operation, one can access information from nodes located further away. For example, $\mathbf{y}^k = \mathbf{S}^k\mathbf{x} = \mathbf{S}(\mathbf{S}^{k-1}\mathbf{x})$ aggregates information from its $k$-hop neighbors (see Fig \ref{fig_aggregation}). Now one can define the spectral $K$-localized graph convolution as :
\begin{equation}\label{eq:z}
    \mathbf{z} = \sum_{k=0}^{K} h_k \mathbf{S}^k \mathbf{x} = \mathbf{H(S)x}
\end{equation}
where $\mathbf{H(S)} = \sum_{k=0}^{\infty} h_k \mathbf{S}^k$ is a linear shift invariant graph filter ~\cite{segarra2017optimal} with coefficients $h_k$. 
Similar to CNNs the output of the GCN is fed into a pointwise non-linear function. Thus, the final form of the graph convolution is given as:
\begin{equation}
\label{eq:finalapproxform}
    \mathbf{z} = \sigma(\mathbf{H(S)x})
\end{equation}
where $\sigma$ is a pointwise non-linearity. A visualization of a GCN can be seen in Fig. \ref{fig_aggregation}.

\subsection{Permutation Equivariance of Graph Convolutional Networks}\label{sec:permutationequivariance}
To control $n$ robots, we propose defining a graph where each robot is a node and robots within $\epsilon$ distance of each other are connected by an edge. The robots are all initialized with random policies and by exploring different actions, they learn what policies best optimize the global reward. Such exploration can change the ordering of the configuration $\mathbf{x}$. This can lead to an explosion in the possible number of graphs that our policies have to learn over. However, ~\cite{gama2019stability} proves a key property for graph convolutional filters. Given a set of permutation matrices : 
\begin{equation}
    \bm{\mathcal{P}} = \{\mathbf{{P}}\in \{0,1\}^{N \times N}\ : \mathbf{{P1=1}},\mathbf{{P}^\top1 =1}\}   
\end{equation}
where the operation $\mathbf{\Bar{P}x}$ permutes the elements of the vector $\mathbf{x}$ then, it can be shown that :
\begin{theorem}\label{theorem:permutationequivariance}
Let graph $\mathcal{G}=(\mathbf{V},\mathbf{E})$ be defined with a graph shift operator $\mathbf{S}$. Further, define $\hat{\mathcal{G}}$ to be the  permuted graph with $\mathbf{\hat{S}} = \mathbf{{P}}^{\top} \mathbf{S} \mathbf{{P}}$ for $\mathbf{{P}} \in \bm{\mathcal{P}}$ and any $\mathbf{x} \in \mathbb{R}^N$ it holds that : 
\begin{equation}
    \mathbf{H}(\hat{\mathbf{S}})\mathbf{P}^{\top}\mathbf{x} = \mathbf{P}^{\top} \mathbf{H(S)x}
\end{equation}
\textit{Proof.} See Appendix 
\end{theorem}
A consequence of Theorem~\ref{theorem:permutationequivariance} is that the output of the graph convolution does not change under reordering of the graph nodes as long as the topology of the graph stays the same. Intuitively, if the graph exhibits several nodes that have the same graph neighborhoods, then the graph convolution filter can be translated to every other node with the same neighborhood. When learning control for a large number of robots, this property helps in reducing dimensionality of the problem. 

\subsection{Formation Flying}
We choose formation flying as a test bed for controlling a large number of robots. In this work, the robots are optimized to produce desired behavior in terms of trajectory following. This behavior can easily be modified from formation flying to other multi-robot objectives such as flocking~\cite{zavlanos2007flocking}, information gathering~\cite{schlotfeldt2018anytime}, collaborative mapping~\cite{ko2003practical} and multi-robot coverage~\cite{choset2001coverage}.  

Consider a two dimensional Euclidean space $\mathbb{R}^2$ with $\mathbf{N}$ homogeneous point mass robots indexed by $n$. In later experiments, we use the robot models defined in the AirSim simulator ~\cite{shah2018airsim} for our experiments. Each robot has a desired goal position $\mathbf{g}_n$ in the euclidean space. Collectively for all robots, all goal locations are denoted $\textbf{g}=[\mathbf{g}_1,\ldots,\mathbf{g}_N]$. At time $t$, the robot's position in the plane is given by $\mathbf{p}_{nt}$. The state representation for robot $n$  that is used by our learning architecture consists only of its own relative position to the goal, i.e $\mathbf{x}_{nt} := \mathbf{p}_{nt} - \mathbf{g}_n$. We choose relative positions to goals as a representation of the robots, in order to maintain permutation invariance. It is important to note here that robot $n$ cannot arbitrarily communicate with any other robot in the swarm. Instead we define a graph $\mathcal{G}$ that encodes pairwise relationships between robots. Robot $n$ can directly communicate only with its one hop neighbors. Communication with neighbors that are further away is possible only indirectly. For each robot, given an action $\mathbf{a}_{nt} \in \mathcal{A}$, the state of the robot evolves according to some stationary dynamics distribution with conditional density $p(\mathbf{x}_{n t+1}|\mathbf{x}_{nt},\mathbf{a}_{nt})$. In this work, we work with continuous control only since it poses a much harder problem and is also more realistic when working with robots. Two conditions are encoded for robots to maintain formation. These are collision avoidance, and waypoint reaching for robots. The necessary and sufficient condition to ensure collision avoidance between robots is given as : 
\begin{equation}
\label{eq:collision_avoidance}
    E_c(\mathbf{p}_{it},\mathbf{p}_{jt}) > \delta, \\ \forall i \neq j \in \{1,\ldots N\}, \forall {t}
\end{equation} 
where $E_c$ is the euclidean distance and $\delta$ is a user-defined minimum distance between robots. Let us also define an assignment matrix $\phi(t) \in \mathbb{R}^{N \times N}$ as :
\begin{equation}
    \phi_{ij}(t) = 
    \begin{cases}
    1, &\text{if }  \forall i=j, E_c(\mathbf{p}_{it},\mathbf{g}_j) \leq \epsilon \\
    0, &\text{otherwise}
    \end{cases}
\end{equation}
where $\epsilon$ is some threshold region of acceptance. The necessary and sufficient condition for all robots to be cover their assigned goals at some time $t=T$ is then:
\begin{equation}
\label{eq:stopping}
    \phi(T)^\top\phi(T) = \textbf{I}_{N}
\end{equation}
where $\textbf{I}$ is the identity matrix. With these definitions in place, we now define the problem statement considered in this paper:  \\
\textbf{Problem 1.} \textit{Given an initial set of robot configurations $\mathbf{x}_0$, a graph $\mathcal{G}$ defining relationships between robots and some goals} $\textbf{g}$\textit{, compute a set of policies $\Pi = [\pi_1,\ldots,\pi_n]$ such that executing actions $\{\mathbf{a}_{1t},\ldots,\mathbf{a}_{Nt}\}= \{\pi_1(\mathbf{a}_{1t}|\mathbf{x}_{1t},\mathcal{G}),\ldots,\pi_N(\mathbf{a}_{Nt}|\mathbf{x}_{Nt},\mathcal{G})\}$ results in a sequence of states that satisfy Eq.\ref{eq:collision_avoidance} and at time $t=T$, satisfy the assignment constraint in Eq.\ref{eq:stopping}.}   
\subsection{Graph Policy Gradients}
We proposing solving the statement in Problem 1 by exploiting the underlying graph structure. Given an initial swarm of $N$ robots, a graph $\mathcal{G}$ can be defined by setting each robot to be a node. Next, an edge is added between nodes, if:
\begin{equation}
\label{eq:graph_node}
    E_c(\mathbf{p}_{i},\mathbf{p}_{j}) \leq \lambda, \\ \forall i \neq j \in \{1,\ldots N\}
\end{equation} 
where $\lambda$ is some user defined threshold to connect two robots.  
It is assumed that at time $t$, the position of robot $n$ (given as $\mathbf{p}_{nt}$) and as a consequence the relative position of robot $n$ (given as $\mathbf{x}_{nt}$) to its own goal is known precisely. The relative positions of all the robots at time $t$ are collected into the vector  $\textbf{x}_t=[\mathbf{x}_{1t},\ldots,\mathbf{x}_{Nt}]$. The graph $\mathcal{G}$ defined earlier, acts as the support for $\textbf{x}_t$.  We do not evolve the graph over time since interchanging homogeneous nodes results in an equivalent graph convolution as discussed in Section \ref{sec:permutationequivariance}. To ensure that the topology stays constant, the number of neighbors for each node is kept fixed, i.e robot $n$ is connected to its say three or four nearest neighbors. We also experiment with a time varying graph, in our simulations and it is observed that for a small number of robots, policies using computed time varying graphs converge slower than fixed graphs (see Fig. \ref{fig:dynamicstatformflying}). However, as the number of robots is increased, policies computed using time varying graphs do not converge whereas policies trained using a fixed graph still converge to desired behavior. One possible reason for this is that when the number of robots is small, the number of possible graphs is also small and it is easier to learn over this small number of graphs. As the number of robots increase, the number of possible graphs increase exponentially leading to a very large number of graphs that the robot needs to learn over. 

To compute the policies, a GCN architecture with $L$ layers is initialized. At each layer according to Eq.\ref{eq:finalapproxform}, the output is given as:
\begin{equation}
\label{eq:layerwiserule}
    \mathbf{z}^{l+1} = \sigma\big(\mathbf{H(S)}\mathbf{z}^l \big)
\end{equation}
where $\sigma$ is a pointwise non-linear function, $\mathbf{z}^0 = \textbf{x}_t$ and $\mathbf{z}^L =\Pi=[\pi_1,\ldots,\pi_N]$. In practice, the final layer outputs are parameters of Gaussian distributions from which actions are sampled. Intuitively at every node, the GCN architecture aggregates information and uses this information to compute policies. 
In order to satisfy the  constraints given in  Eq.\ref{eq:collision_avoidance} and Eq.\ref{eq:stopping}, we formulate a \textit{centralized} reward structure of the form:
\begin{equation}
\label{eq:rewardstruc}
    r(t) =
    \begin{cases}
   -\beta, &\text{if }  \text{any collisions},\\
   -\sum_{i}^N E_c(\mathbf{p}_{it},\mathbf{g}_i) &\text{ } \text{otherwise} \\
    \end{cases}
\end{equation}
Each robot receives the same reward an attempts to learn a policy that best optimizes this reward. It is assumed that the policies for the robots are independent. Thus, the overall loss function for all the robots as given in Eq.\ref{eq:overallloss}
\begin{equation}
J = \sum_{n=1}^N \max_{\theta}  \mathbb{E}_{\Pi}\bigg[\sum_{t}^T r_t\bigg]  \enspace
\end{equation}
where $\theta$ now represents the filter weights of the GCN for $\Pi$. Consider a trajectory  $\tau=(\mathbf{x_0},\mathbf{a_0},\ldots,\mathbf{x_T},\mathbf{a_T})$. 
Since the reward along a trajectory is the same for all robots and all robot policies are assumed independent, using direct differentiation the policy gradient is given as :
\begin{equation}
\label{eq:policygradient}
    \nabla_{\theta}J = \mathbb{E}_{\tau \sim (\pi_1,\ldots,\pi_N)}\Bigg[\Big(\sum_{t=1}^T\nabla_{\theta} \log[\pi_1(\mathbf{a}_{1t}|\mathbf{x}_{1t},\mathcal{G})\ldots\pi_N(\mathbf{a}_{Nt}|\mathbf{x}_{Nt},\mathcal{G})]\Big) \Big(\sum_{t=1}^T r_t \Big) \Bigg] 
\end{equation}
For the full algorithm, please see the appendix. The weights $\theta$, are then updated using any variant of stochastic gradient descent. We call this algorithm Graph Policy Gradients (GPG). In the next section we demonstrate how GPG is able to learn meaningful policies even as the number of robots increase and is also able to transfer filters learned from a small number of robots to a larger number of robots. 
\section{Experiments}
To investigate the performance of GPG, we design experiments to answer three key questions :
\begin{itemize}
    \item Can GPG learn policies that achieve desired behavior as the number of robots increase?
    \item How well do the graph filters learned by GPG transfer to a large number of robots?
    \item Does GPG work with more complex dynamics and controls?
\end{itemize}
\subsection{Training for Point Mass Formation Flying with GPG}\label{sec:pointmassexps}
\begin{wrapfigure}{l}{0.3\textwidth}
  \begin{center}
    \includegraphics[scale=0.3]{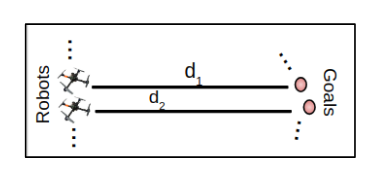}
  \end{center}
  \caption{\textbf{Formation Flying.}}\label{fig:formflying}
\end{wrapfigure}
We first look to establish if GPG can train for formation flying for simple point mass dynamics as the number of robots are increased. The state of every robot $n$ is its absolute position to its goal and the action $a_{nt}$ at time $t$ for robot $n$ is the change in $x$ and $y$ position in the plane. Such a setting necessitates communication between robots. One could argue that the robots simply have to learn to take actions such that $\textbf{x}_{nt}$ tends to zero. However, consider the example in Fig~\ref{fig:formflying}. When the goals are reasonably far away $d_1\approx d_2$ and each robot can take actions that could cause it to collide with its neighbors. Each robot must communicate with its neighbors who in turn must communicate with theirs in order to achieve collision free trajectories that take robots to their assigned goals.

\begin{wrapfigure}{r}{0.4\textwidth}
  \begin{center}
    \includegraphics[scale=0.23]{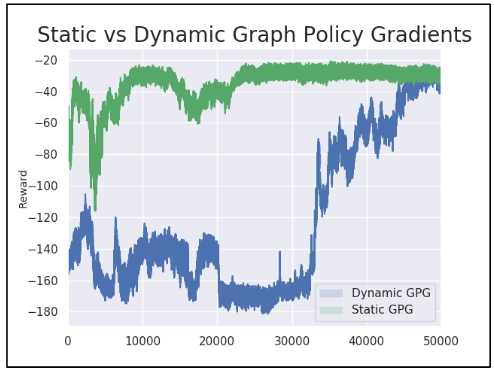}
  \end{center}
  \caption{\textbf{10 robot Formation Flying.} Using a static graph during training increases the sample efficiency of GPG. A dynamic graph, i.e a graph that evolves over time as the robots move in space takes longer to converge.}\label{fig:dynamicstatformflying}
\end{wrapfigure}
\begin{figure}[t!]
  \centering
  \includegraphics[width=\textwidth]{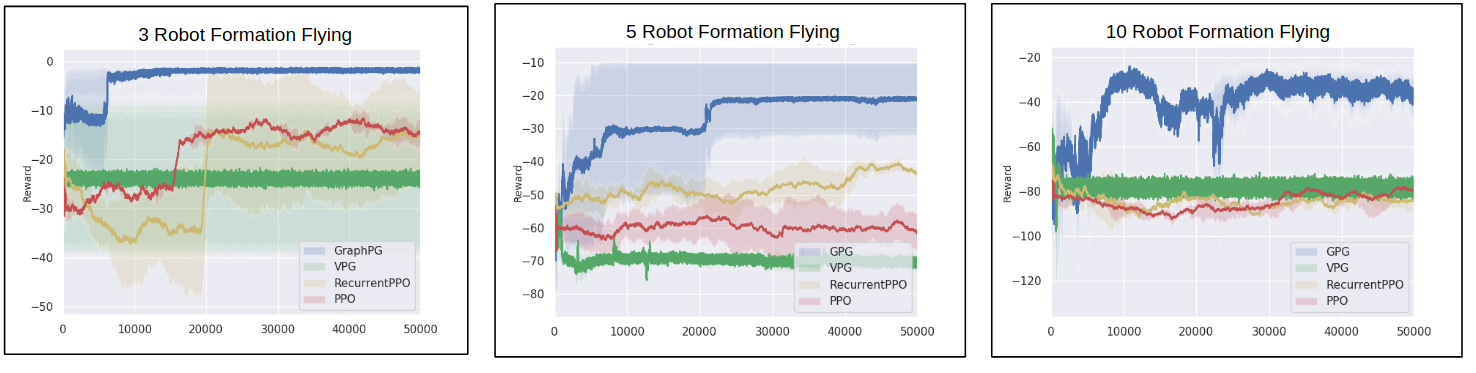}
  \caption{\textbf{Training for Formation Flight.} Point mass robots are trained for formation flight. The reward is a centralized reward. Each curve is produced by running three independent runs of the algorithm. Darker line represents mean and shaded area represents mean $\pm$ standard deviation of mean.\label{fig:curvesfig}}
  \vspace{-1cm}
\end{figure}

To establish relevant baselines, we choose vanilla policy gradients (VPG) that uses the same policy gradient hyperparameters as GPG but differs from GPG in that it uses a 2 layer fully connected network for policy paramterization. We also compare with Proximal Policy Optimization (PPO) ~\cite{schulman2017proximal} a state of the art on policy method for learning continuous policies. 
In this paper, we employ a batched version of PPO for faster computation~\cite{hafner2017agents}. The PPO baseline also employs fully connected networks. Another choice of baseline used in this paper PPO with a recurrent network architecture since the problem on hand requires communication between agents and shared memory is one possible mechanism. The training curves for our experiments with 3,5 and 10 robots can be seen in Fig~\ref{fig:curvesfig}. We observe right away that GPG is able to converge in all three cases. VPG does not produce any meaningful behavior and both variants of PPO only produce partial results.  
In the experiments above, we maintain a static graph $\mathcal{G}$ while training in light of the permutation equivariance property of GCNs.  While static graphs are used during training to speed up the training process, during testing, the graphs are dynamic since the graph convolution yields the same result. We also experiment with a dynamic graph $\mathcal{G}_t$ that evolves as the robots move in space. As hypothesized and guided by the intuition of Theorem~\ref{theorem:permutationequivariance}, using dynamic graphs increases sample complexity of the problem whereas the static graph converges faster. This can be seen in Fig~\ref{fig:dynamicstatformflying}.
\subsection{Zero Shot Policy Transfer for Formation Flying}\label{sec:zeroshotexps}
It can be observed from Fig~\ref{fig:curvesfig} that while GPG is able to converge as the number of robots increase, the number of samples required also increases. Training GPG for ten robots alone takes over nine hours on a 12GB NVIDIA 2080Ti GPU whereas we wish to learn policies for hundreds of robots. One possible solution is to realize that the filters trained extract local features and the same filter can be used when working with many more robots. To demonstrate this, we train a filter with a small number of robots (depending on the formation we are interested in) and use the same filter weights without any gradient updates on the larger swarm.
\begin{figure}[b!]
  \centering
  \includegraphics[width=\textwidth]{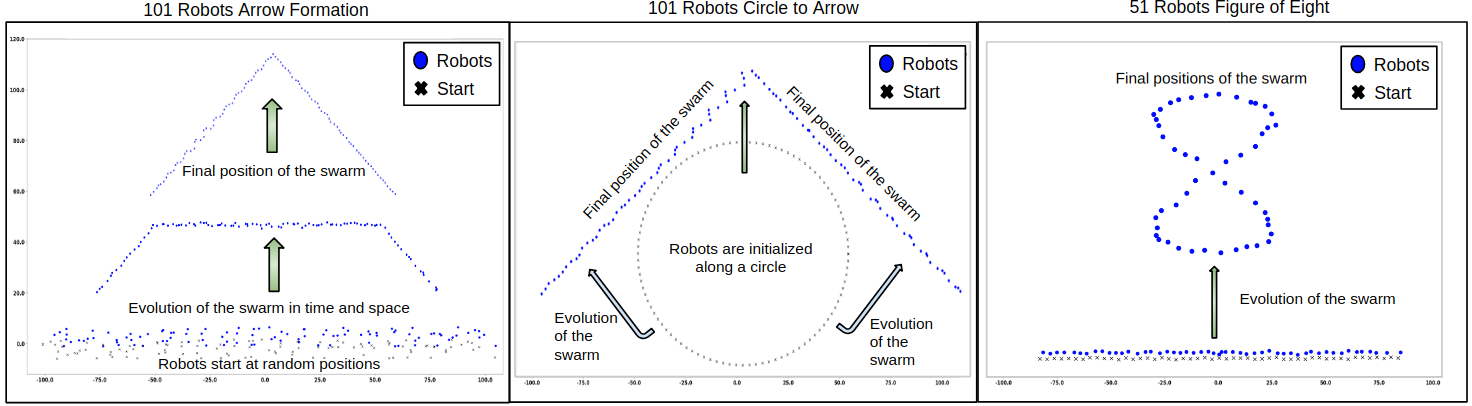}
  \caption{\textbf{Zero Shot Transfer to Large Number of Robots.} (Left) Policies are trained for three robots to reach goals that are a small distance away. Robots are randomly initialized in a rectangular region and must reach goals much further than those in the training set. (Center) Robots are initialized on a circle and must execute policies such that the resulting shape is an arrowhead. (Right) Policies trained on swarms of five robots are transferred over to form a figure of eight. The choice of 101/51 robots is arbitrary and is not a limiting threshold.  \label{fig:formationfig}}
  \vspace{-0.2cm}
\end{figure}
Another point to note is that when training for the smaller swarm, we only train for goals that are a small distance away to reduce the number of samples (it is easier to discover goals that are close by than it is to discover goals much further away) required for training. However, during execution these policies are able to converge even when the goals are hundred units away. A snapshot of some of the formations we can produce using this zero shot transfer can be seen in Fig. \ref{fig:formationfig}.  
In Fig. \ref{fig:formationfig} (left) and (center), policies are trained on three robots and utilize only 1-hop neighbor information. The figure of eight formation necessitates more communication especially when robots at the edges cross over to form the figure of eight. In Fig. \ref{fig:formationfig} (right) the filters are learned by training with five robots with each robot utilizing 3 hop neighbor information.

With these results, it can be concluded that GPG is capable of learning complex behaviors for a small number of robots. These behaviors can then be transferred to larger more complex swarms. To the best of our knowledge there exists no other method to learn continuous control policies that can achieve similar results for so many robots. 
\subsection{Complex dynamics and control} 
In the experiments considered above, actions for robot $n$ are simply the change in position in the plane. Further, it is assumed that full control of robots can be achieved and at every step robots have zero velocities. These assumptions are infeasible in the real world. Thus, we also conduct experiments to test the efficacy of GPG on more realistic dynamics and control. 
\begin{wrapfigure}{r}{0.4\textwidth}
  \begin{center}
    \includegraphics[scale=0.3]{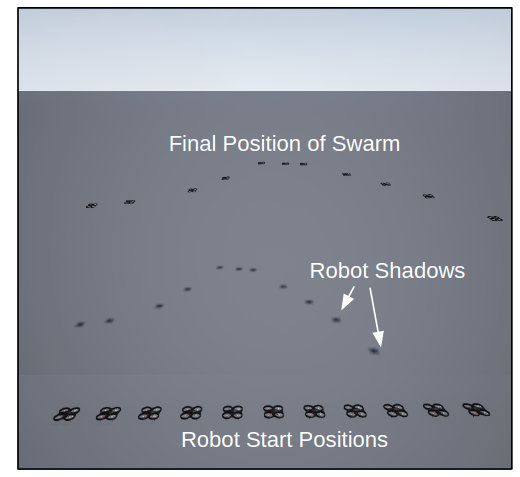}
  \end{center}
  \caption{\textbf{11 robot Arrow Head Formation.} The robots are spawned at ground and are manually controlled for takeoff. Once the robots are at a desired height, control is handed over to GPG. (see Appendix for detailed figures)}\label{fig:airsimflying}
\end{wrapfigure}
To simulate this, we use the AirSim simulator introduced in ~\cite{shah2018airsim}. Robots are now assumed to have finite mass and inertia and obey single integrator dynamics. The state for each robot used by GPG is now given as $\mathbf{x}_{nt}:=[(\mathbf{p}_{nt}-\mathbf{g}_n),\Dot{\mathbf{p}}_{nt}]$ where $\Dot{\mathbf{p}}_{nt}$ is the velocity at time $t$ for robot $n$.  Thus, the state for each robot is not just its relative position but also includes its current velocity. The actions now represent change in velocity in the plane.  Similar to the point mass experiments, policies are first learned for a small number of robots in AirSim. In addition to the non simplistic dynamics,  the difficulty of training policies in AirSim is further compounded by the fact that AirSim is a real time simulator and produces an order of magnitued fewer training points as compared to the point-mass simulation over a fixed period of time. Thus, it is even more imperative when working with such a simulator to be able to learn policies for a small number of robots that require fewer samples and then be able to transfer these policies to a larger number of robots for which direct training would otherwise be infeasible. GPG being a model free algorithm faces no additional difficulty in adapting to the single integrator dynamics. A snapshot of results produced using GPG in AirSim can be seen in Fig~\ref{fig:dynamicstatformflying}. In this case, the baselines used in Section~\ref{sec:pointmassexps} are unable to learn reasonable behaviors within 50K episodes. This can be attributed to the fact that the observations in AirSim are noisier and the baselines most likely need a lot more samples to converge.

\section{Related Work}
In the recent past, there has been an immense amount of interest in learning policies for a large collection of agents. Literature in multi-agent RL has argued for a centralized training decentralized execution scheme~\cite{lowe2017multi,foerster2018counterfactual} where each agent has its own policy network but during train time, it has access to an additional critic network that co-ordinates information among all agents. Naturally, such a scheme is not scalable as the input size of the critic network grows as the number of robots increase in addition to the inherent complexity of training hundreds of separate policies simultaneously. Another line of thinking in this field, has been the idea to approximate the policies of all the agents using meta-learning ~\cite{parisotto2019concurrent,khan2018scalable}. These works while impressive, still run into the problem of having to execute rollouts for a large number of agents to learn a meaningful meta-representation. Closest to our work, perhaps is the work of ~\cite{jiang2018graph} where the authors propose using CNNs on an encoded graph representation of the agents. In contrast to our work, the policies learned are discrete and the convolutions do not fully exploit the local graph structure. 
\section{Conclusion}
The problem of learning policies that map state representation to control commands for a large number of robots is immensely challenging, even if the robots are homogeneous. In this work, we propose using the underlying graph structure as additional information. To exploit information from the graph we propose using GCNs to parametrize the policies. Using GCNs we are able to learn local filters defined on the graph that extract local information. The additional benefit of using these local feature extractors is that they can be used to alleviate the problem of needing a large number of rollouts as the number of robots increase as the number of robots increase as demonstrated by
the experiments in Section \ref{sec:zeroshotexps}. It might even be possible to add other sensor representations such as camera images or lidar instead of requiring exact positions of the robots in space. These could be first processed using standard
architectures such as CNNs and then be fed into GPG to learn policies for a large swarm of robots
directly from on board sensors. We leave this for future work.


\clearpage
\acknowledgments{We gratefully acknowledge the support of ARL grant ARL DCIST CRA W911NF-17-2-0181. This work was supported in part by the Semiconductor Research Corporation (SRC), DARPA, NVIDIA and Intel. 
}


\bibliography{example}  
\clearpage
\appendix
\section{Permutation Equivariance}
The proof for Theorem 1 was first given in ~\cite{gama2019stability}. We reiterate it here due to its importance to this body of work.

Given a set of permutation matrices : 
\begin{equation}
    \bm{\mathcal{P}} = \{\mathbf{{P}}\in \{0,1\}^{N \times N}\ : \mathbf{{P1=1}},\mathbf{{P}^\top1 =1}\}   
\end{equation}
where the operation $\mathbf{\Bar{P}x}$ permutes the elements of the vector $\mathbf{x}$ then, it can be shown that :

\textit{\textbf{Theorem 1}}
Let graph $\mathcal{G}=(\mathbf{V},\mathbf{E})$ be defined with a graph shift operator $\mathbf{S}$. Further, define $\hat{\mathcal{G}}$ to be the  permuted graph with $\mathbf{\hat{S}} = \mathbf{{P}}^{\top} \mathbf{S} \mathbf{{P}}$ for $\mathbf{{P}} \in \bm{\mathcal{P}}$ and any $\mathbf{x} \in \mathbb{R}^N$ it holds that : 
\begin{equation}
    \mathbf{H}(\hat{\mathbf{S}})\mathbf{P}^{\top}\mathbf{x} = \mathbf{P}^{\top} \mathbf{H(S)x}
\end{equation}

\textit{Proof for Theorem 1} \\
Given that $\mathbf{{P}}$ is a permutation matrix. This implies $\mathbf{{P}}$ is also an orthogonal matrix. This implies $\mathbf{P}^{\top}\mathbf{{P}}=\mathbf{{P}}\mathbf{{P}}^\top=\mathbf{I}$. Thus, 
\begin{equation}
    \hat{\mathbf{S}}^k = \mathbf{{P}}^\top \mathbf{S}^k \mathbf{{P}} 
\end{equation}
Then, 
\begin{equation}
    \mathbf{H}(\hat{\mathbf{S}})= \sum_{k=0}^\infty h_k \hat{\mathbf{S}}^k = \sum_{k=0} h_k (\mathbf{{P}}^\top \mathbf{S}^k \mathbf{{P}}) = \mathbf{P}^{\top} \mathbf{H(S)} \mathbf{P}
\end{equation}
Finally, using $\mathbf{P}\mathbf{P}^{\top}=\mathbf{I}$
\begin{equation}
    \mathbf{H}(\hat{\mathbf{S}})\mathbf{P}^\top \mathbf{x} = \mathbf{P}^\top \mathbf{H(S)}\mathbf{P}\mathbf{P}^\top \mathbf{x} = \mathbf{P}^\top \mathbf{H(s)} \mathbf{x}
\end{equation}
Hence, proved. 
\section{Experiments in AirSim}
For the experiments in AirSim, we assume the robots to obey single integrator dynamics.
\begin{wrapfigure}{r}{0.3\textwidth}
  \begin{center}
    \includegraphics[scale=0.26]{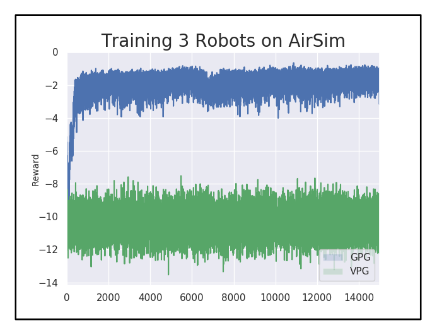}
  \end{center}
  \caption{\textbf{Training 3 Robots in AirSim.}}\label{fig:airsimtraining}
\end{wrapfigure}
Robot $n$'s position in the plane at time $t$ is given as $\mathbf{p}_{nt}$ and its velocity at time $t$ is given as $\Dot{\mathbf{p}}_{nt}$. The action $\mathbf{a}_{nt}$ chosen by the robot gives the change in velocities. The state of robot $n$ then evolves according to :
\begin{equation}
     \begin{bmatrix}
           \mathbf{p}_{nt+1} \\
           \Dot{\mathbf{p}}_{nt+1} \\
         \end{bmatrix}
         = \begin{bmatrix}
           \mathbf{I} \hspace{0.3cm} T_s\mathbf{I}\\
           0 \hspace{0.3cm} \mathbf{I}\\
         \end{bmatrix}
              \begin{bmatrix}
           \mathbf{p}_{nt} \\
           \Dot{\mathbf{p}}_{nt} \\ 
         \end{bmatrix}
         + 
         \begin{bmatrix}
           0 \\
           0.1 T_s \mathbf{I} \\
         \end{bmatrix}
         \mathbf{a}_{nt}
\end{equation}
where $T_s$ is sampling time. 
To speed up training, the velocities are clipped in a range of $[-1,1]$m/s. The robots are initialized along a straight line with a fixed distance between them. An external controller is used to arm the robots and for takeoff. Once all the robots are at a fixed height, control is handed over to GPG. Further, when the robots reach their goals, they do not enter hover mode. GPG outputs small velocities for the robots and as a result when robots are close to their goals, they tend to oscillate around the goal point. 
\begin{figure}[t!]
  \centering
  \includegraphics[width=\textwidth]{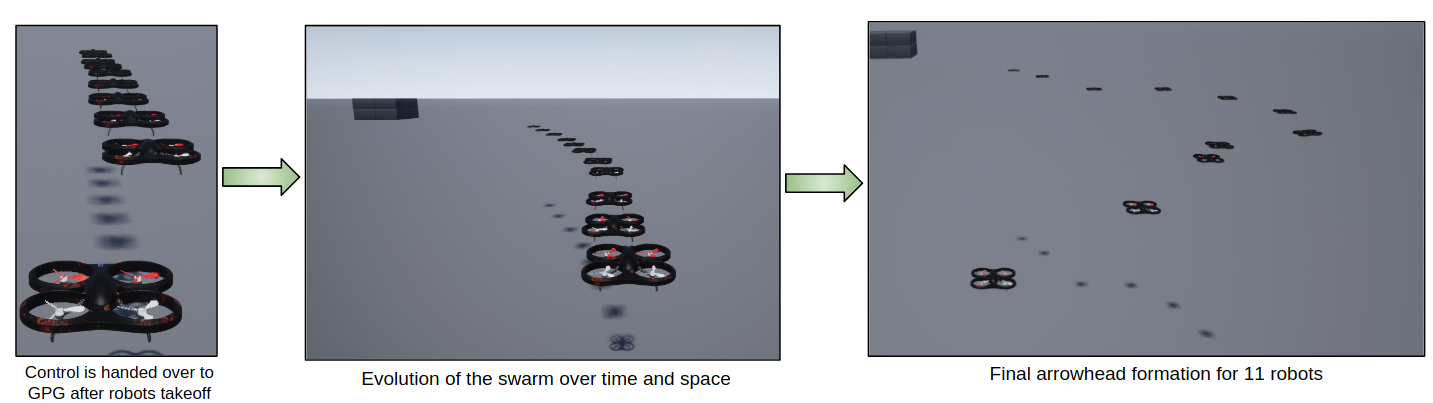}
  \caption{\textbf{Arrowhead Formation Flying for 11 robots in AirSim.} \label{fig:airsimformationfig}}
\end{figure}
\section{Hyperparameters}
For the point mass experiments, when learning one-hop neighbor information, we use a 2 layer GCN with a tanh nonlinearity. The input layer consists of 16 hidden units and is fed the state representation $\textbf{x}_t$. The output of this layer is then fed into a $\mu$ layer and a $\sigma$ layer both of which have 16 hidden units. Actions for the robots are sampled using these parameters. We use the Deep Graph Library (DGL) to represent our GCNs. The learning rate is set to $1$e$-3$ and uses the Adam optimizer. For the VPG baseline, the learning rate and the optimizer remain the same. The network is parametrized by fully connected layers where the input to the first layer is the state representation and consists of 64 hidden units followed by a ReLU non linearity. The output of this first layer is then fed into separate $\mu$ and $\sigma$ layers each consisting of 64 hidden units and ReLU activation functions. We invite the reader to take a look at the code attached with this paper for more details. 
\section{Graph Policy Gradients}
Algorithm \ref{algo_gpg} summarizes the full graph policy gradients algorithm. 
\begin{algorithm}[h!]  %
Initialize $N$ robots and policies $\Pi = [\pi_1,\ldots,\pi_N]$ parametrized by a GCN whose weights are given by $\theta$ and graph $\mathcal{G}$ with graph shift operator $\mathbf{S}$
\begin{algorithmic}[1]
\While {True}
\For {time $t=[0,\ldots, T]$}
\For {robots $n=[1,\ldots,N]$}
\State Record each robots state $\mathbf{x}_t$
\State Aggregate information at each robot $n$ from its neighbors $\mathfrak{B}_n$ according to Eq. \ref{eq:graph_signal}
\begin{align*}
    y_n =  \sum_{j=n,j\in \mathfrak{B}_n}s_{nj}\mathbf{x}_{nt}
\end{align*}
\EndFor
\State Collect $\textbf{x}_t = [\mathbf{x}_{1t},\ldots,\mathbf{x}_{Nt}]$
\State Compute $K$ localized graph convolution $\mathbf{z}_t$ as given in Eq.\ref{eq:z} 
\begin{align*}
        \mathbf{z}_t = \sum_{k=0}^{K} h_k \mathbf{S}^k \textbf{x}_t = \mathbf{H(S)\textbf{x}_t}
\end{align*}
\State Stack $L$ graph convolution layers such that $\mathbf{z}_t^0 = \textbf{x}_t$ and $\mathbf{z}_t^L=\Pi$
\begin{align*}
    \mathbf{z}_t^{l+1} = \sigma\big(\mathbf{H(S)}\mathbf{z}_t^l \big)
\end{align*}
\State Sample $\textbf{a}_t= \{\mathbf{a}_{1t},\ldots,\mathbf{a}_{Nt}\} := \{\pi_1(\mathbf{a}_{1t}|\mathbf{x}_{1t},\mathcal{G}),\ldots,\pi_N(\mathbf{a}_{Nt}|\mathbf{x}_{Nt},\mathcal{G})\}$
\State Execute $\textbf{a}_t$ and collect reward over all robots $R_t$=$\sum_{n=1}^N r(t)$
\EndFor
\State Record trajectory $\tau=\big[(\mathbf{x}_0,\mathbf{a}_0,R_0), \ldots,(\mathbf{x}_T,\mathbf{a}_T,R_T$)\big].
\State Compute the graph policy gradients as given in Eq \ref{eq:policygradient} and update weights $\theta$
\EndWhile
\end{algorithmic}
\caption{Graph Policy Gradients.}\label{algo_gpg}
\end{algorithm}

\end{document}